\documentclass{article}

\PassOptionsToPackage{numbers}{natbib}
\usepackage[final]{nips_2017}

\usepackage[utf8]{inputenc} 
\usepackage[T1]{fontenc}    
\usepackage{hyperref}       
\usepackage{url}            
\usepackage{booktabs}       
\usepackage{amsfonts}       
\usepackage{nicefrac}       
\usepackage{microtype}      

\usepackage{amsmath}
\usepackage{amssymb}
\usepackage{amsthm}
\usepackage{graphicx}
\usepackage{algorithm}
\usepackage{algorithmic}
\usepackage{comment}
\usepackage{caption}
\usepackage{subcaption}

\newtheorem{theorem}{Theorem}
\newtheorem{proposition}[theorem]{Proposition}

\title{Distributed Solution of Large-Scale Linear Systems \\ via Accelerated Projection-Based Consensus}

\author{
  Navid Azizan-Ruhi~$^1$, Farshad Lahouti~$^1$, Salman Avestimehr~$^2$, Babak Hassibi~$^1$\\
  \\
  $^1$~California Institute of Technology, CA 91125\\
  $^2$~University of Southern California, CA 90089\\
  \texttt{\{azizan,lahouti,hassibi\}@caltech.edu}, \texttt{avestimehr@ee.usc.edu}
}

\begin{document}

\maketitle

\begin{abstract}
Solving a large-scale system of linear equations is a key step at the heart of many algorithms in machine learning, scientific computing, and beyond. When the problem dimension is large, computational and/or memory constraints make it desirable, or even necessary, to perform the task in a distributed fashion. In this paper, we consider a common scenario in which a taskmaster intends to solve a large-scale system of linear equations by distributing subsets of the equations among a number of computing machines/cores. We propose an accelerated distributed consensus algorithm, in which at each iteration every machine updates its solution by adding a scaled version of the projection of an error signal onto the nullspace of its system of equations, and where the taskmaster conducts an averaging over the solutions with momentum. The convergence behavior of the proposed algorithm is analyzed in detail and analytically shown to compare favorably with the convergence rate of alternative distributed methods, namely distributed gradient descent, distributed versions of Nesterov's accelerated gradient descent and heavy-ball method, the block Cimmino method, and ADMM. On randomly chosen linear systems, as well as on real-world data sets, the proposed method offers significant speed-up relative to all the aforementioned methods. Finally, our analysis suggests a novel variation of the distributed heavy-ball method, which employs a particular distributed preconditioning, and which achieves the same theoretical convergence rate as the proposed consensus-based method.
\end{abstract}


\section{Introduction}

With the advent of big data, many analytical tasks of interest rely on distributed computations over multiple processing cores or machines.
This is either due to the inherent complexity of the problem, in terms of computation and/or memory, or due to the nature of the data sets themselves that may already be dispersed across machines. Most algorithms in the literature have been designed to run in a sequential fashion, as a result of which in many cases their distributed counterparts have yet to be devised. 

Many sophisticated algorithms in machine learning and data analysis are composed of a number of basic computations (e.g.,  matrix algebra, solving equations, etc). For these computations to run efficiently in a distributed setting, we are required to address a number of technical questions: (a) What computation should each worker carry out? (b) What is the communication architecture and what messages should be communicated between the processors? (c) How does the distributed implementation fare in terms of computational complexity? and (d) What is the rate of convergence in the case of iterative algorithms?

\begin{figure}[t]
\begin{center}
\centerline{\includegraphics[width=0.45\columnwidth]{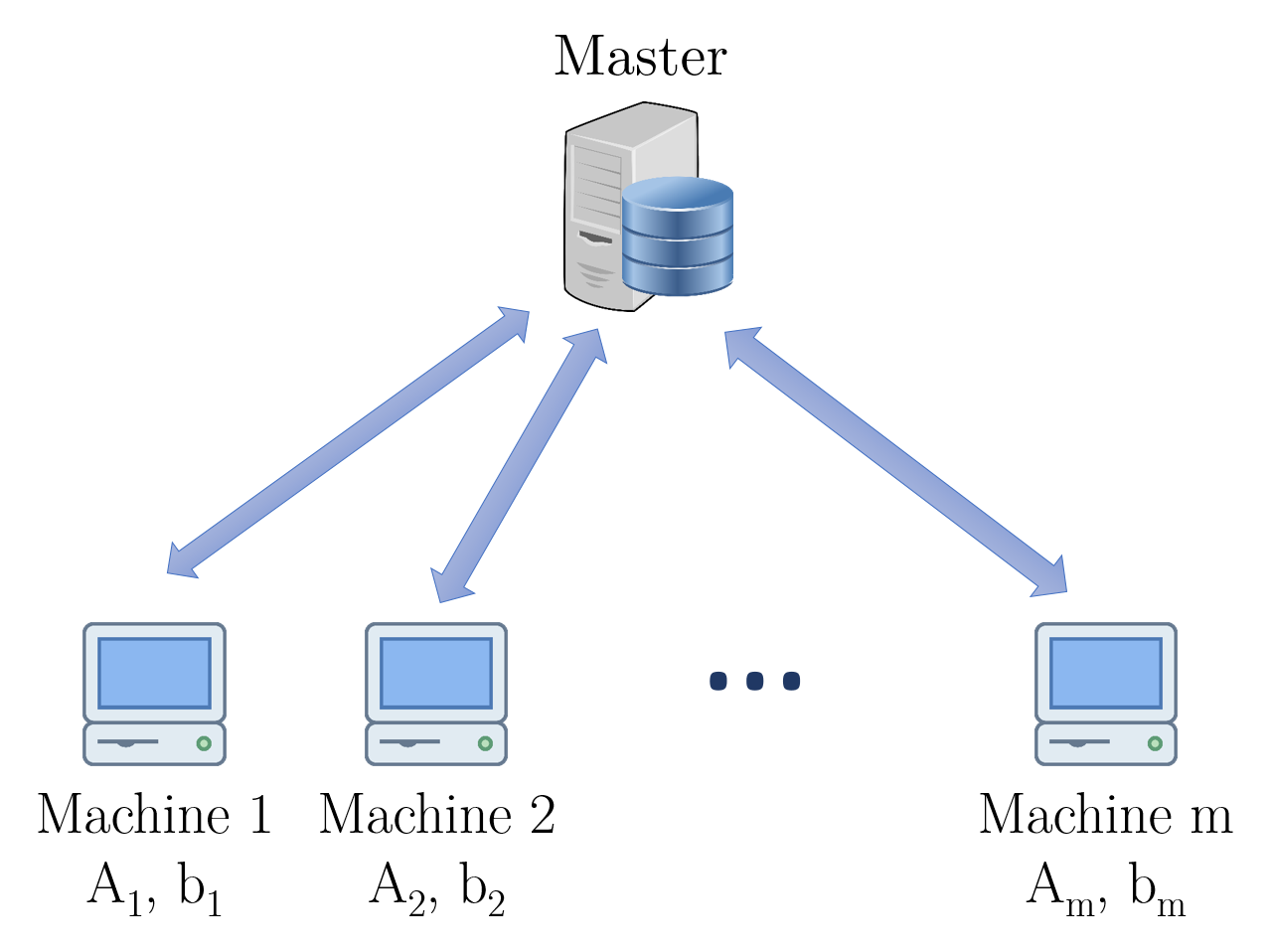}}
\caption{Schematic representation of the taskmaster and the $m$ machines. Each machine $i$ has only a subset of the equations, i.e. $[A_i, b_i]$.}
\label{fig:1}
\end{center}
\end{figure}

In this paper, we focus on solving a system of linear equations in a distributed fashion, which is one of the most fundamental problems in numerical computation, and lies at the heart of many algorithms in 
engineering and the sciences. In particular, we consider the setting in which a taskmaster intends to solve a large-scale system of equations with the help of a set of computing machines/cores (Figure~\ref{fig:1}).

This problem can in general be cast as an optimization problem with a cost function that is separable in the data (but not in the variables) \footnote{Solving a system of linear equations, $Ax=b$, can be set up as the optimization problem $\min_x\|Ax-b\|^2=\min_x \sum_i \|(Ax)_i-b_i\|^2$.}. Hence, there are general approaches to construct distributed algorithms for this problem, such as distributed versions of gradient descent and its variants (e.g. Nesterov's accelerated gradient \cite{nesterov1983method}, heavy-ball method \cite{polyak1964some}, etc.), where each machine computes the partial gradient corresponding to a term in the cost and the taskmaster then aggregates the partial gradients by summing them, as well as the so-called Alternating Direction Method of Multipliers (ADMM) and its variants \cite{boyd2011distributed}. Among others, some recent approaches for Distributed Gradient Descent (DGD) have been presented and analyzed in \cite{zinkevich2010parallelized}, \cite{recht2011hogwild} and \cite{yuan2016convergence}, and also coding techniques for robust DGD in the presence of failures and straggler machines have been studied in \cite{
LMA16Unify, tandon2016gradient}. ADMM has been widely used \cite{He2012,Deng2012,zhang2014asynchronous} for solving various convex optimization problems in a distributed way, and in particular for consensus optimization \cite{Mota2013,shi2014linear,Layla16}, which is the relevant one for the type of separation that we have here.

In addition to the optimization-based methods, there are a few distributed algorithms designed specifically for solving systems of linear equations. The most famous one of these is what is known as the block Cimmino method \cite{duff2015augmented,sloboda1991projection,arioli1992block}, which is a block row-projection method \cite{bramley1992row}, and is in a way a distributed implementation of the Kaczmarz method \cite{kaczmarz1937angenaherte}. Another algorithm has been recently proposed in \cite{liu2013asynchronous,mou2015distributed}, where a consensus-based scheme is used to solve a system of linear equations over a network of autonomous agents. Our algorithm bears some resemblance to all of these methods, but as it will be explained in detail, it has much faster convergence than any of them.

Our main contribution is the design and analysis of a fast distributed consensus-based algorithm to solve large-scale systems of linear equations. 
More specifically, we develop a methodology in which the taskmaster assigns a subset of equations to each of the machines and invokes a distributed consensus-based algorithm to obtain the solution to the original problem in an iterative manner. At each iteration, each machine updates its solution by projecting an error signal onto the nullspace of its corresponding system of equations and taking a weighted step in that direction. The taskmaster then conducts a memory-augmented averaging on the new solutions provided by the machines. We prove that the algorithm has linear convergence, and we quantify its convergence rate and complexity.
Compared to the Kaczmarz/Cimmino-type methods \cite{duff2015augmented,sloboda1991projection,arioli1992block} and the consensus scheme proposed in \cite{liu2013asynchronous,mou2015distributed}, our method is significantly accelerated, due to the momentum incorporated in both projection and averaging steps. For this reason, we refer to our method as \emph{Accelerated Projection-Based Consensus} (APC). 
We provide a complete analysis of the convergence rate of APC (Section~\ref{sec:APC}), as well as a detailed comparison with all the other distributed methods mentioned above (Section~\ref{sec:other}).  
Also by empirical evaluations over both randomly chosen linear systems and real-world data sets, we demonstrate that the proposed algorithm offers significant speed-ups relative to the other distributed methods (Section~\ref{sec:experiments}).
Finally, as a further implication of our results, we propose a novel distributed preconditioning method (Section~\ref{sec:pre}), which can be used to improve the convergence rate of distributed gradient-based methods and match that of the APC.

\section{The Setup}
We consider the problem of solving a large-scale system of linear equations
\begin{equation}\label{problem}
Ax=b,
\end{equation}
where $A\in\mathbb{R}^{N\times n}$, $x\in\mathbb{R}^n$ and $b\in\mathbb{R}^N$. While we will generally take $N\geq n$, we will assume that the system has a unique solution. 
For this reason, we will most often consider the square case ($N=n$).

As mentioned before, for large-scale problems (when $N,n\gg 1$), it is highly desirable, or even necessary, to solve the problem in a distributed fashion.
Assuming we have $m$ machines, the equations can be partitioned so that each machine gets a disjoint subset of them. In other words, we can write \eqref{problem} as
$$\begin{bmatrix}A_1\\A_2\\ \vdots\\ A_m\end{bmatrix}x=\begin{bmatrix}b_1\\b_2\\ \vdots\\ b_m\end{bmatrix}, $$
where each machine $i$ receives $[A_i,b_i]$. In some applications, the data may already be stored on different machines in such a fashion.
For the sake of simplicity, we assume that $m$ divides $N$, and that the equations are distributed evenly among the machines, so that each machine gets $p=\frac{N}{m}$ equations. Therefore $A_i\in\mathbb{R}^{p\times n}$ and $b_i\in\mathbb{R}^{p}$ for every $i=1,\dots m$.
It is helpful to think of $p$ as being relatively small compared to $n$. In fact, each machine has a system of equations which is highly under-determined.

\section{Accelerated Projection-Based Consensus}\label{sec:APC}

\subsection{The Algorithm}
Each machine $i$ can certainly find a solution (among infinitely many) to its own highly under-determined system of equations $A_ix=b_i$, with simply $O(p^3)$ computations. We denote this initial solution by $x_i(0)$. Clearly adding any vector in the right nullspace of $A_i$ to $x_i(0)$ will yield another viable solution. The challenge is to find vectors in the nullspaces of each of the $A_i$'s in such a way that all the solutions for different machines coincide.

At each iteration $t$, the master provides the machines with an estimate of the solution, denoted by $\bar{x}(t)$. Each machine then updates its value $x_i(t)$ by projecting its difference from the estimate onto the nullspace, and taking a weighted step in that direction (which behaves as a ``momentum''). Mathematically
$$x_i(t+1) =x_i(t)+ \gamma P_i(\bar{x}(t)-x_i(t)) ,$$
where $P_i=I-A_i^T(A_iA_i^T)^{-1}A_i$ is the projection matrix onto the nullspace of $A_i$ (It is easy to check that $A_iP_i=0$ and $P_i^2=P_i$).

Although this might have some resemblance to the block Cimmino method because of the projections, as it will be explained, the convergence time of the APC method is much better (by a square root) than that of the block Cimmino method. Moreover, it turns out that the block Cimmino method is in fact a special case of APC for $\gamma=1$.

The update rule of $x_i(t+1)$ described above can be also thought of as the solution to an optimization problem with two terms, the distance from the global estimate $\bar{x}(t)$ and the distance from the previous solution $x_i(t)$. In other words, one can show that
\begin{equation*}
\begin{aligned}
x_i(t+1)=\arg& \underset{x_i}{\min}
& & \|x_i-\bar{x}(t)\|^2+\frac{1-\gamma}{\gamma}\|x_i-x_i(t)\|^2 \\
& \text{ s.t.}
& & A_ix_i=b_i
\end{aligned}
\end{equation*}
The second term in the objective is what distinguishes this method from the block Cimmino method. If one sets $\gamma$ equal to $1$ (which, as it will be shown in Section~\ref{sec:cimmino}, is the reduction to the block Cimmino method), the second term  disappears altogether, and the update no longer depends on $x_i(t)$. As we will show, this has a dramatic impact on the convergence rate.

After each iteration, the master collects the updated values $x_i(t+1)$ to form a new estimate $\bar{x}(t+1)$. A plausible choice for this is to simply take the average of the values as the new estimate, i.e.,
$$\bar{x}(t+1)=\frac{1}{m}\sum_{i=1}^m x_i(t+1) .$$
This update works, and is what appears both in ADMM and in the consensus method of \cite{liu2013asynchronous,mou2015distributed}. But it turns out that it is extremely slow. Instead, we take an affine combination of the average and the previous estimate as
$$\bar{x}(t+1)=\frac{\eta}{m}\sum_{i=1}^m x_i(t+1) + (1-\eta)\bar{x}(t),$$
which introduces a one-step memory, and again behaves as a momentum. 

The resulting update rule is therefore
\begin{subequations}\label{eq:cons_gamma}
\begin{align}
x_i(t+1)&=x_i(t)+\gamma P_i(\bar{x}(t)-x_i(t)),\quad i=1,\dots,m ,\label{eq:cons1_gamma}\\
\bar{x}(t+1)&=\frac{\eta}{m}\sum_{i=1}^m x_i(t+1) + (1-\eta)\bar{x}(t) ,\label{eq:cons2_gamma}
\end{align}
\end{subequations}
which leads to Algorithm~\ref{alg:APC}.

\begin{algorithm}[tb]
   \caption{APC: Accelerated Projection-based Consensus (For solving $Ax=b$ distributedly)}
   \label{alg:APC}
\begin{algorithmic}
   \STATE {\bfseries Input:} data $[A_i,b_i]$ for each machine $i=1,\dots m$, parameters $\eta,\gamma$
   \STATE {\bfseries Initialization:} at each machine $i$ find a solution $x_i(0)$ (among infinitely many) to $A_ix=b_i$.
   \FOR{$t=1$ {\bfseries to} $T$}
   \FOR{each machine $i$ {\bfseries parallel}}
   \STATE{$x_i(t+1)\gets x_i(t)+\gamma P_i(\bar{x}(t)-x_i(t))$}
   \ENDFOR
   \STATE{at the master: $\bar{x}(t+1)\gets \frac{\eta}{m}\sum_{i=1}^m x_i(t+1) + (1-\eta)\bar{x}(t)$}
   \ENDFOR
\end{algorithmic}
\end{algorithm}

\subsection{Convergence Analysis}
We analyze the convergence of the proposed algorithm and prove that it has linear convergence with no additional assumption imposed. We also derive the rate of convergence explicitly.

Let us define the matrix $X\in\mathbb{R}^{n\times n}$ as
\begin{equation}
X\triangleq\frac{1}{m}\sum_{i=1}^m A_i^T(A_iA_i^T)^{-1}A_i .
\end{equation}
Note that $\frac{1}{m}\sum_{i=1}^m P_i = I - X$. As it will become clear soon, the condition number of this matrix determines the behavior of the algorithm. Since the eigenvalues of the projection matrix $P_i$ are all $0$ and $1$, for every $i$, the eigenvalues of $X$ are all between $0$ and $1$. Denoting the eigenvalues of $X$ by $\mu_i$, we have:
\begin{equation}
0\leq\mu_{\min}\triangleq\mu_n\leq\dots\leq\mu_1\triangleq\mu_{\max}\leq 1 .
\end{equation}
Let us define complex quadratic polynomials $p_i(\lambda)$ characterized by $\gamma$ and $\eta$ as
\begin{equation}
p_i(\lambda;\gamma,\eta)\triangleq\lambda^2+\left(-\eta\gamma(1-\mu_i)+\gamma-1+\eta-1\right)\lambda+(\gamma-1)(\eta-1)
\end{equation}
for $i=1,\dots,n$. Further, define set $S$ as the collection of pairs $\gamma\in[0,2]$ and $\eta\in\mathbb{R}$ for which the largest magnitude solution of $p_i(\lambda)=0$ among every $i$ is less than $1$, i.e.
\begin{equation}
S=\{(\gamma,\eta)\in[0,2]\times\mathbb{R} \mid \text{ roots of }p_i\text{ have magnitude less than }1 \text{ for all } i\}.
\end{equation}
The following result summarizes the convergence behavior of the proposed algorithm.

\begin{theorem}\label{thm:APC}
Algorithm~\ref{alg:APC} converges to the true solution as fast as $\rho^t$ converges to $0$, as $t\to\infty$, for some $\rho\in(0,1)$, if and only if $(\gamma,\eta)\in S$. Furthermore, the optimal rate of convergence is
\begin{equation}
\rho
=\frac{\sqrt{\kappa(X)}-1}{\sqrt{\kappa(X)}+1} \approx 1-\frac{2}{\sqrt{\kappa(X)}},
\end{equation}
where $\kappa(X)=\frac{\mu_{\max}}{\mu_{\min}}$ is the condition number of $X$, and the optimal parameters $(\gamma^*,\eta^*)$ are the solution to the following equations
$$\begin{cases}\mu_{\max}\eta\gamma=(1+\sqrt{(\gamma-1)(\eta-1)})^2,\\\mu_{\min}\eta\gamma=(1-\sqrt{(\gamma-1)(\eta-1)})^2.\end{cases}$$
\end{theorem}
For proof see the supplementary material.

\subsection{Computational Complexity and Numerical Stability}
In addition to the convergence rate, or equivalently the number of iterations until convergence, one needs to consider the computational complexity per iteration.

At each iteration, since $P_i=I_n-A_i^T(A_iA_i^T)^{-1}A_i$, and $A_i$ is $p\times n$, each machine has to do the following two matrix-vector multiplications: (1) $A_i(x_i(t)-\bar{x}(t))$, which takes $pn$ scalar multiplications, and (2) $\left(A_i^T(A_iA_i^T)^{-1}\right)$ times the vector from the previous step (e.g. using QR factorization), which takes another $np$ operations.
Thus the overall computational complexity of each iteration is $2pn$.

Finally, we should mention that the computation done at each machine during each iteration is essentially a projection, which has condition number one and is as numerically stable as a matrix vector multiplication can be. 

\section{Comparison with Related Methods}\label{sec:other}

\subsection{Distributed Gradient Descent (DGD)}
As mentioned earlier, \eqref{problem} can also be viewed as an optimization problem of the form
$$\underset{x}{\text{minimize }} \|Ax-b\|^2 ,$$
and since the objective is separable in the data, i.e. $\|Ax-b\|^2=\sum_{i=1}^m \|A_ix-b_i\|^2$, generic distributed optimization methods such as distributed gradient descent apply well to the problem.

The regular or full gradient descent has the update rule
$x(t+1)=x(t)-\alpha A^T(Ax(t)-b)$,
where $\alpha>0$ is the step size or learning rate.
The distributed version of gradient descent is one in which each machine $i$ has only a subset of the equations $[A_i,b_i]$, and computes its own part of the gradient, which is $A_i^T(A_ix(t)-b_i)$. The updates are then collectively done as:
\begin{equation}
x(t+1)=x(t)-\alpha \sum_{i=1}^m A_i^T(A_ix(t)-b_i) .
\end{equation}

One can show that this also has linear convergence, and the rate of convergence is
\begin{equation}\label{eq:rho(M')}
\rho
= \frac{\kappa(A^TA)-1}{\kappa(A^TA)+1}\approx 1-\frac{2}{\kappa(A^TA)}.
\end{equation}

We should mention that since each machine needs to compute $A_i^T(A_ix(t)-b_i)$ at each iteration $t$, the computational complexity per iteration is $2pn$, which is identical to that of APC.

\subsection{Distributed Nesterov's Accelerated Gradient Descent (D-NAG)}
A popular variant of gradient descent is Nesterov's accelerated gradient descent \cite{nesterov1983method}, which has a memory term, and works as follows:
\begin{subequations}
\begin{align}
y(t+1) &= x(t)-\alpha \sum_{i=1}^m A_i^T(A_ix(t)-b_i) ,\\
x(t+1) &= (1+\beta)y(t+1)-\beta y(t) .
\end{align}
\end{subequations}
One can show \cite{lessard2016analysis} that the optimal convergence rate of this method is 
\begin{equation}
\rho= 1-\frac{2}{\sqrt{3\kappa(A^TA)+1}} ,
\end{equation}
which is improved over the regular distributed gradient descent.

\subsection{Distributed Heavy-Ball Method (D-HBM)}
The heavy-ball method \cite{polyak1964some}, otherwise known as the gradient descent with momentum, is another accelerated variant of gradient descent as follows:
\begin{subequations}
\begin{align}
z(t+1)= &\beta z(t)+\sum_{i=1}^m A_i^T(A_ix(t)-b_i) ,\\
x(t+1)= &x(t)-\alpha z(t+1) .
\end{align}
\end{subequations}
It can be shown \cite{lessard2016analysis} that the optimal rate of convergence of this method is 
\begin{equation}
\rho
= \frac{\sqrt{\kappa(A^TA)}-1}{\sqrt{\kappa(A^TA)}+1} \approx 1-\frac{2}{\sqrt{\kappa(A^TA)}},
\end{equation}
which is further improved over DGD and D-NAG, and is similar to, but not the same as, APC. The difference is that the condition number of $A^TA = \sum_{i=1}^mA_i^TA_i$ is replaced with the condition number of $X = \sum_{i=1}^mA_i^T\left(A_iA_i^T\right)^{-1}A_i$ in APC. Given its structure as the sum of projection matrices, one may speculate that $X$ has a much better condition number than $A^TA$. Indeed, our experiments with random, as well as real, data sets suggest that this is the case and that the condition number of $X$ is often significantly better.

\begin{table}[t]
  \caption{A summary of the convergence rates of different methods. The smaller the convergence rate is, the faster is the method. DGD: Distributed Gradient Descent, D-NAG: Distributed Nesterov's Accelerated Gradient Descent, D-HBM Distributed Heavy-Ball Method, Consensus: Standard Projection-Based Consensus, B-Cimmino: Block Cimmino Method, APC: Accelerated Projection-based Consensus.}
  \label{table:summary}
  \centering
  \begin{tabular}{cccccc}
    \toprule
    DGD     & D-NAG   & D-HBM  & Consensus & B-Cimmino & APC (proposed)\\
    \midrule
    $1-\frac{2}{\kappa(A^TA)}$  & $1-\frac{2}{\sqrt{3\kappa(A^TA)+1}}$ & $1-\frac{2}{\sqrt{\kappa(A^TA)}}$ & $1-\mu_{\min}(X)$ & $1-\frac{2}{\kappa(X)}$ & $1-\frac{2}{\sqrt{\kappa(X)}}$\\
    \bottomrule
  \end{tabular}
\end{table}

\subsection{Alternating Direction Method of Multipliers (ADMM)}\label{sec:ADMM}
Alternating Direction Method of Multipliers (more specifically, consensus ADMM \cite{shi2014linear,boyd2011distributed}), is another generic method for solving optimization problems with separable cost function $f(x)=\sum_{i=1}^m f_i(x)$ distributedly, by defining additional local variables.
Each machine $i$ holds local variables $x_i(t)\in\mathbb{R}^n$ and $y_i(t)\in\mathbb{R}^n$, and the master's value is $\bar{x}(t)\in\mathbb{R}^n$, for any time $t$. 
For $f_i(x)=\frac{1}{2}\|A_ix-b_i\|^2$, 
the update rule of ADMM simplifies to
\begin{subequations}
\begin{align}
x_i(t+1) &=(A_i^TA_i+\xi I_n)^{-1}(A_i^Tb_i-y_i(t)+\xi \bar{x}(t)),\quad i=1,\dots,m\label{eq:admm_1}\\
\bar{x}(t+1) &= \frac{1}{m}\sum_{i=1}^m x_i(t+1)\label{eq:admm_2}\\
y_i(t+1) &= y_i(t) + \xi(x_i(t+1)-\bar{x}(t+1)),\quad i=1,\dots,m
\end{align}
\end{subequations}

It turns out that this method is very slow (and often unstable) in its native form for the application in hand. One can check that when system \eqref{problem} has a solution all the $y_i$ variables converge to zero in steady state. Therefore setting $y_i$'s to zero can speed up the convergence significantly. We use this modified version in Section~\ref{sec:experiments}, to compare with. 

We should also note that the computational complexity of ADMM is $O(pn)$ per iteration (the inverse is computed using matrix inversion lemma), which is again the same as that of gradient-type methods and APC.

\subsection{Block Cimmino Method}\label{sec:cimmino}
The Block Cimmino method \cite{duff2015augmented,sloboda1991projection,arioli1992block}, which is a parallel method specifically for solving linear systems of equations, is perhaps the closest algorithm in spirit to APC. It is, in a way, a distributed implementation of the so-called Kaczmarz method \cite{kaczmarz1937angenaherte}. The convergence of the Cimmino method is slower by an order in comparison with APC (its convergence time is the square of that of APC), and it turns out that APC includes this method as a special case when $\gamma=1$.

The block Cimmino method is the following:
\begin{subequations}
\begin{align}
&r_i(t) = A_i^+(b_i-A_i\bar{x}(t)),\quad i=1,\dots,m\label{eq:cimmino_1}\\
&\bar{x}(t+1) = \bar{x}(t)+\nu \sum_{i=1}^m r_i(t)\label{eq:cimmino_2} ,
\end{align}
\end{subequations}
where $A_i^+=A_i^T(A_iA_i^T)^{-1}$ is the pseudoinverse of $A_i$.

\begin{proposition}\label{prop}
The APC method (Algorithm~\ref{alg:APC}) includes the block Cimmino method as a special case for $\gamma=1$.
\end{proposition}
The proof is provided in the supplementary material.

It is not hard to show that optimal rate of convergence of the Cimmino method is
\begin{equation}
\rho=\frac{\kappa(X)-1}{\kappa(X)+1} \approx 1-\frac{2}{\kappa(X)},
\end{equation}
which is by an order worse than that of APC ($1-\frac{2}{\sqrt{\kappa(X)}}$).

A summary of the optimal convergence rates of all the methods is provided in Table~\ref{table:summary}

\section{Experimental Results}\label{sec:experiments}

\begin{table}[tbh]
\caption{A comparison between the optimal convergence time $T$ ($=\frac{1}{-\log\rho}$) of different methods on real and synthetic examples. Boldface values show the smallest convergence time.\\ QC324: Model of $H_2^+$ in an Electromagnetic Field. ORSIRR 1: Oil Reservoir Simulation. ASH608: Original Harwell sparse matrix test collection.}
\label{table1}
\begin{center}
\begin{small}
\begin{sc}
\begin{tabular}{lccccccr}
\toprule
 & DGD & D-NAG & D-HBM & M-ADMM & B-Cimmino & APC\\
\midrule
\begin{tabular}{@{}l@{}}QC324 \\ ($324\times 324$) \end{tabular} & $1.22\times 10^7$ & $4.28\times 10^3$ & $2.47\times 10^3$ & $1.07\times 10^7$ & $3.10\times 10^5$ & $\bf3.93\times 10^2$\\
\midrule
\begin{tabular}{@{}l@{}}ORSIRR 1\\ ($1030\times 1030$)\end{tabular} & $2.98\times 10^9$ & $6.68\times 10^4$ & $3.86\times 10^4$ & $2.08\times 10^8$ & $2.69\times 10^7$ & $\bf3.67\times 10^3$\\
\midrule
\begin{tabular}{@{}l@{}}ASH608 \\  ($608\times 188$)\end{tabular}  & $5.67\times 10^0$ & $2.43\times 10^0$ & $1.64\times 10^0$ & $1.28\times 10^1$ & $4.98\times 10^0$ & $\bf1.53\times 10^0$\\
\midrule
\begin{tabular}{@{}l@{}}Standard\\ Gaussian \\ ($500\times 500$)\end{tabular}   & $1.76\times 10^7$ & $5.14\times 10^3$ & $2.97\times 10^3$ & $1.20\times 10^6$ & $1.46\times 10^7$ & $\bf2.70\times 10^3$\\
\midrule
\begin{tabular}{@{}l@{}}Nonzero-Mean\\Gaussian\\ ($500\times 500$)\end{tabular}   & $2.22\times 10^{10}$ & $1.82\times 10^5$ & $1.05\times 10^5$ & $8.62\times 10^8$ & $9.29\times 10^8$ & $\bf2.16\times 10^4$\\
\midrule
\begin{tabular}{@{}l@{}}Standard Tall\\ Gaussian\\ ($1000\times 500$)\end{tabular}    & $1.58\times 10^1$ & $4.37\times 10^0$ & $2.78\times 10^0$ & $4.49\times 10^1$ & $1.13\times 10^1$ & $\bf2.34\times 10^0$\\
\bottomrule
\end{tabular}
\end{sc}
\end{small}
\end{center}
\end{table}

\begin{figure*}[t]
    \centering
    \begin{subfigure}[b]{0.49\textwidth}
        \includegraphics[width=\textwidth]{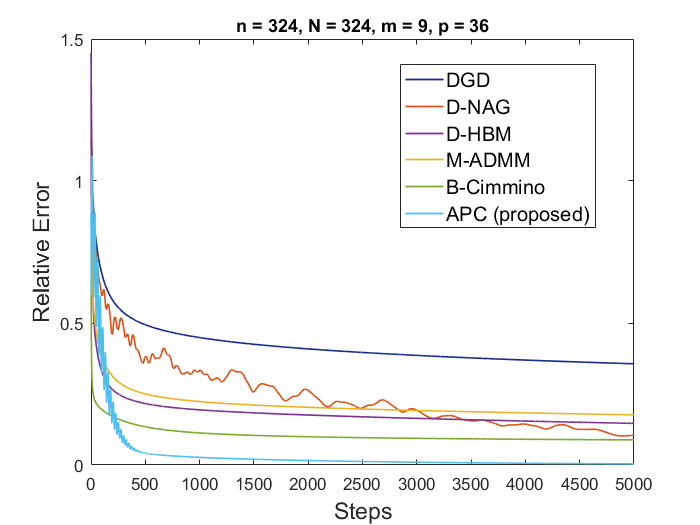}
        \label{fig:qc324}
    \end{subfigure}
    \begin{subfigure}[b]{0.49\textwidth}
        \includegraphics[width=\textwidth]{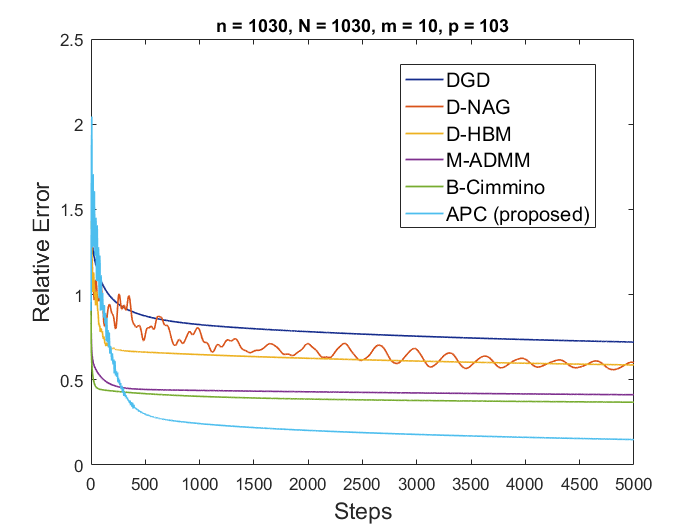}
        \label{fig:orsirr1}
    \end{subfigure}
    \caption{The decay of the error for different distributed algorithms, on two real problems from Matrix Market \cite{MatrixMarket} (QC324: Model of $H_2^+$ in an Electromagnetic Field, and ORSIRR 1: Oil reservoir simulation). $n=\#$ of variables, $N=\#$ of equations, $m=\#$ of workers, $p=\#$ of equations per worker.}\label{fig:decay}
\end{figure*}

In this section, we evaluate the proposed method (APC) by comparing it with the other distributed methods discussed throughout the paper, namely DGD, D-NAG, D-HBM, modified ADMM, and block Cimmino methods.
We use randomly-generated problems as well as real-world ones form the National Institute of Standards and Technology repository, \emph{Matrix Market} \cite{MatrixMarket}.

We first compare the rates of convergence of the algorithms $\rho$, which is the spectral radius of the iteration matrix. To distinguish the differences, it is easier to compare the convergence times, which is defined as $T=\frac{1}{-\log\rho}$ ($\approx\frac{1}{1-\rho}$). We tune the parameters in all of the methods to their optimal values, to make the comparison between the methods fair. Also as mentioned before, all the algorithms have the same per-iteration computation and communication complexity. Table~\ref{table1} shows the values of the convergence times for a number of synthetic and real-world problems with different sizes. It can be seen that APC has a much faster convergence, often by orders of magnitude. As expected from the analysis, the APC's closest competitor is the distributed heavy-ball method. Notably, in randomly-generated problems, when the mean is not zero, the gap is much larger.

To further verify the performance of the proposed algorithm, we also run all the algorithms on multiple problems, and observe the actual decay of the error.  Fig.~\ref{fig:decay} shows the relative error (the distance from the true solution, divided by the true solution, in $\ell_2$ norm) for all the methods, on two examples from the repository. Again, to make the comparison fair, all the methods have been tuned to their optimal parameters. As one can see, APC outperforms the other methods by a wide margin, which is consistent with the order-of-magnitude differences in the convergence times of Table~\ref{table1}. We should also remark that initialization does not seem to affect the convergence behavior of our algorithm.

\section{A Distributed Preconditioning to Improve Gradient-Based Methods}\label{sec:pre}
The noticeable similarity between the optimal convergence rate of APC ($\frac{\sqrt{\kappa(X)}-1}{\sqrt{\kappa(X)}+1}$) and that of D-HBM ($\frac{\sqrt{\kappa(A^TA)}-1}{\sqrt{\kappa(A^TA)}+1}$) suggests that there might be a connection between the two. It turns out that there is, and we propose a distributed preconditioning for D-HBM, which makes it achieve the same convergence rate as APC. The algorithm works as follows.

Prior to starting the iterative process, each machine $i$ can premultiply its own set of equations $A_ix=b_i$ by $(A_iA_i^T)^{-1/2}$, which can be done in parallel (locally) with $O(p^2n)$ operations. This transforms the global system of equations $Ax=b$ to a new one $Cx=d$, where
$$C=\begin{bmatrix}(A_1A_1^T)^{-1/2}A_1\\ \vdots\\ (A_mA_m^T)^{-1/2}A_m\end{bmatrix}, \quad\text{ and }\quad
d=\begin{bmatrix}(A_1A_1^T)^{-1/2}b_1\\ \vdots\\ (A_mA_m^T)^{-1/2}b_m\end{bmatrix} .$$

The new system can then be solved using distributed heavy-ball method, which will achieve the same rate of convergence as APC, i.e. $\frac{\sqrt{\kappa}-1}{\sqrt{\kappa}+1}$ where $\kappa=\kappa(C^TC)=\kappa(X)$.

\section{Conclusion}
We considered the problem of solving a large-scale system of linear equations by a taskmaster with the help of a number of computing machines/cores, in a distributed way. We proposed an accelerated projection-based consensus algorithm for this problem, and fully analyzed its convergence rate. 
Analytical and experimental comparisons with the other known distributed methods confirm significantly faster convergence of the proposed scheme. Finally, our analysis suggested a novel distributed preconditioning for improving the convergence of the distributed heavy-ball method to achieve the same theoretical performance as the proposed consensus-based method.


\bibliography{references}
\bibliographystyle{plain}

\newpage
\begin{center}
{\huge Supplementary Material}
\newline
\end{center}
\appendix

\section{Proof of the Main Result}
Here we provide the proof of Theorem~\ref{thm:APC}.
\begin{proof}
Let $x^*$ be the unique solution of $Ax=b$. To make the analysis easier, we define error vectors with respect to $x^*$ as $e_i(t)=x_i(t)-x^*$ for all $i=1\dots m$, and $\bar{e}(t)=\bar{x}(t)-x^*$, and work with these vectors. Using this notation, Eq.~\eqref{eq:cons1_gamma} can be rewritten as
\begin{equation*}
e_i(t+1)=e_i(t)+\gamma P_i(\bar{e}(t)-e_i(t)),\quad i=1,\dots,m .
\end{equation*}
Note that both $x^*$ and $x_i(t)$ are solutions to $A_ix=b_i$. Therefore, their difference, which is $e_i(t)$, is in the nullspace of $A_i$, and it remains unchanged under projection onto the nullspace. As a result, $P_ie_i(t)=e_i(t)$, and we have
\begin{equation}\label{eq:e_i(t+1)_gamma}
e_i(t+1) = (1-\gamma)e_i(t)+\gamma P_i\bar{e}(t),\quad i=1,\dots,m .
\end{equation}

Similarly, the recursion \eqref{eq:cons2_gamma} can be expressed as
\begin{equation*}
\bar{e}(t+1)=\frac{\eta}{m}\sum_{i=1}^m e_i(t+1)+(1-\eta) \bar{e}(t),
\end{equation*}
which using \eqref{eq:e_i(t+1)_gamma} becomes
\begin{align}
\bar{e}(t+1)&=\frac{\eta}{m}\sum_{i=1}^m \left((1-\gamma)e_i(t)+\gamma P_i\bar{e}(t)\right)+(1-\eta) \bar{e}(t)\notag\\
&=\frac{\eta(1-\gamma)}{m}\sum_{i=1}^m e_i(t)+\left(\frac{\eta\gamma}{m}\sum_{i=1}^m P_i + (1-\eta)I_n\right)\bar{e}(t) \label{eq:bar{e}(t+1)}.
\end{align}

It is relatively easy to check that in the steady state, the recursions~\eqref{eq:e_i(t+1)_gamma}, \eqref{eq:bar{e}(t+1)} become 
$$
\begin{cases}
P_i\bar{e}(\infty)=e_i(\infty),\quad i=1,\dots,m \\
\bar{e}(\infty)=\frac{1}{m}\sum_{i=1}^m P_i\bar{e}(\infty)
\end{cases}
$$
which because of $\frac{1}{m}\sum_{i=1}^m P_i=I-\frac{1}{m}\sum_{i=1}^m A_i^T(A_iA_i^T)^{-1}A_i =I-X$, implies $\bar{e}(\infty)=e_1(\infty)=\dots=e_m(\infty)=0$, if $\mu_{\min}\neq 0$.

Now let us stack up all the $m$ vectors $e_i$ along with the average $\bar{e}$ together, as a vector $e(t)^T=[e_1(t)^T, e_2(t)^T,\dots,e_m(t)^T,\bar{e}(t)^T]\in\mathbb{R}^{(m+1)n}$. The update rule can be expressed as:
\begin{equation}\label{eq:matrix_recursion}
\begin{bmatrix}e_1(t+1)\\ \vdots\\e_m(t+1)\\ \bar{e}(t+1)\end{bmatrix}=
\begin{bmatrix}
(1-\gamma)I_{mn}& \gamma\begin{bmatrix}P_1\\ \vdots\\ P_m\end{bmatrix}\\
\frac{\eta(1-\gamma)}{m}\begin{bmatrix}I_n\dots I_n\end{bmatrix}& M
\end{bmatrix}
\begin{bmatrix}e_1(t)\\ \vdots\\e_m(t)\\ \bar{e}(t)\end{bmatrix} ,
\end{equation}
where
$M=\frac{\eta\gamma}{m}\sum_{i=1}^m P_i+(1-\eta)I_n$.

The convergence rate of the algorithm is determined by the spectral radius (largest magnitude eigenvalue) of the $(m+1)n\times(m+1)n$ block matrix in \eqref{eq:matrix_recursion}. The eigenvalues $\lambda_i$ of this matrix are indeed the solutions to the following characteristic equation.
$$\det
\begin{bmatrix}
(1-\gamma-\lambda)I_{mn}& \gamma\begin{bmatrix}P_1\\ \vdots\\ P_m\end{bmatrix}\\
\frac{\eta(1-\gamma)}{m}\begin{bmatrix}I_n\dots I_n\end{bmatrix}& \frac{\eta\gamma}{m}\sum_{i=1}^m P_i+(1-\eta-\lambda)I_n
\end{bmatrix}=0 .$$
Using the Schur complement and properties of determinant, the characteristic equation can be simplified as follows.
\begin{align*}
0 &=(1-\gamma-\lambda)^{mn}\det\left(\frac{\eta\gamma}{m}\sum_{i=1}^m P_i+(1-\eta-\lambda)I_n-\frac{\eta(1-\gamma)\gamma}{(1-\gamma-\lambda)m}\sum_{i=1}^m P_i\right)\\
&=(1-\gamma-\lambda)^{mn}\det\left(\frac{\eta\gamma}{m}(1-\frac{1-\gamma}{1-\gamma-\lambda})\sum_{i=1}^m P_i+(1-\eta-\lambda)I_n\right)\\
&=(1-\gamma-\lambda)^{mn}\det\left(\frac{-\eta\gamma\lambda}{(1-\gamma-\lambda)m}\sum_{i=1}^m P_i+(1-\eta-\lambda)I_n\right)\\
&=(1-\gamma-\lambda)^{(m-1)n}\det\left(-\eta\gamma\lambda\frac{\sum_{i=1}^m P_i}{m}+(1-\gamma-\lambda)(1-\eta-\lambda)I_n\right) .
\end{align*}
Therefore, there are $(m-1)n$ eigenvalues equal to $1-\gamma$, and the remaining $2n$ eigenvalues are the solutions to
\begin{align*}
0 &=\det\left(-\eta\gamma\lambda(I-X)+(1-\gamma-\lambda)(1-\eta-\lambda)I\right)\\
&=\det\left(\eta\gamma\lambda X+\left((1-\gamma-\lambda)(1-\eta-\lambda)-\eta\gamma\lambda\right)I\right) .
\end{align*}
Whenever we have dropped the subscript of the identity matrix, it is of size $n$.

Recall that the eigenvalues of $X$ are denoted by $\mu_i,\ i=1,\dots,n$. Therefore, the eigenvalues of $\eta\gamma\lambda X+\left((1-\gamma-\lambda)(1-\eta-\lambda)-\eta\gamma\lambda\right)I$ are $\eta\gamma\lambda \mu_i+(1-\gamma-\lambda)(1-\eta-\lambda)-\eta\gamma\lambda,\ i=1,\dots,n$. The above determinant can then be written as the product of the eigenvalues of the matrix inside it, as
$$0=\prod_{i=1}^n \eta\gamma\lambda \mu_i+(1-\gamma-\lambda)(1-\eta-\lambda)-\eta\gamma\lambda .$$

Therefore, there are two eigenvalues $\lambda_{i,1}, \lambda_{i,2}$ as the solution to the quadratic equation
$$\lambda^2+(-\eta\gamma(1-\mu_i)+\gamma-1+\eta-1)\lambda+(\gamma-1)(\eta-1)=0$$
for every $i=1,\dots,n$, which will constitute the $2n$ eigenvalues. When all these eigenvalues, along with $1-\gamma$, are less than $1$, the error converges to zero as $\rho^t$, with $\rho$ being the largest magnitude eigenvalue (spectral radius). Therefore, Algorithm~\ref{alg:APC} converges to the true solution $x^*$ as fast as $\rho^t$ converges to $0$, as $t\to\infty$, if and only if $(\gamma,\eta)\in S$.

The optimal rate of convergence is achieved when the spectral radius is minimum. For that to happen, all the above eigenvalues should be complex and have magnitude $|\lambda_{i,1}|=|\lambda_{i,2}|=\sqrt{(\gamma-1)(\eta-1)}=\rho$. It implies that we should have
$$(\gamma+\eta-\eta\gamma(1-\mu_i)-2)^2\leq 4(\gamma-1)(\eta-1),\quad \forall i,$$
or equivalently
$$-2\sqrt{(\gamma-1)(\eta-1)}\leq\gamma+\eta-\eta\gamma(1-\mu_i)\leq 2\sqrt{(\gamma-1)(\eta-1)},\quad \forall i.$$
The expression in the middle is an increasing function of $\mu_i$, and therefore for the above bounds to hold, it is enough for the lower bound to hold for the $\mu_{\min}$ and the upper bound to hold for $\mu_{\max}$, i.e.
$$\begin{cases}\gamma+\eta-\eta\gamma(1-\mu_{\max})-2=2\sqrt{(\gamma-1)(\eta-1)}\\2+\eta\gamma(1-\mu_{\min})-\gamma-\eta=2\sqrt{(\gamma-1)(\eta-1)} \end{cases} ,$$
which can be massaged and expressed as
$$\begin{cases}\mu_{\max}\eta\gamma=(1+\sqrt{(\gamma-1)(\eta-1)})^2=(1+\rho)^2\\ \mu_{\min}\eta\gamma=(1-\sqrt{(\gamma-1)(\eta-1)})^2=(1-\rho)^2\end{cases}.$$
Dividing the above two equations implies $\kappa(X)=\frac{(1+\rho)^2}{(1-\rho)^2}$, which results in the optimal rate of convergence being
$$\rho=\frac{\sqrt{\kappa(X)}-1}{\sqrt{\kappa(X)}+1}$$

\end{proof}

\section{Proof of Proposition~\ref{prop}}
We prove that the block Cimmino method is a special case of the proposed algorithm for $\gamma=1$.
\begin{proof}
When $\gamma=1$, Eq.~\eqref{eq:cons1_gamma} becomes
\begin{align*}
x_i(t+1) &=x_i(t)-P_i\left(x_i(t)-\bar{x}(t)\right)\\
&=x_i(t)-\left(I-A_i^T(A_iA_i^T)^{-1}A_i\right)\left(x_i(t)-\bar{x}(t)\right)\\
&=\bar{x}(t)+A_i^T(A_iA_i^T)^{-1}A_i(x_i-\bar{x}(t))\\
&=\bar{x}(t)+A_i^T(A_iA_i^T)^{-1}(b_i-A_i\bar{x}(t))
\end{align*}
In the last equation, we used the fact that $x_i$ is always a solution to $A_ix=b_i$. Notice that the above equation is no longer an ``update'' in the usual sense, i.e., $x_i(t+1)$ does not depend on $x_i(t)$ directly. This can be further simplified using the pseudoinverse of $A_i$, $A_i^+=A_i^T(A_iA_i^T)^{-1}$  as
$$x_i(t+1)=\bar{x}(t)+A_i^+(b_i-A_i\bar{x}(t)) .$$
It is then easy to see from the Cimmino's equation \eqref{eq:cimmino_1} that $$r_i(t)=x_i(t+1)-\bar{x}(t) .$$
Therefore, the update \eqref{eq:cimmino_2} can be expressed as
\begin{align*}
\bar{x}(t+1)&=\bar{x}(t)+\nu \sum_{i=1}^m r_i(t)\\
&=\bar{x}(t)+\nu\sum_{i=1}^m\left(x_i(t+1)-\bar{x}(t)\right)\\
&=(1-m\nu )\bar{x}(t)+\nu\sum_{i=1}^m x_i(t+1),
\end{align*}
which is nothing but the same update rule as in \eqref{eq:cons2_gamma} with $\eta=m\nu$.
\end{proof}

\end{document}